\documentclass[letterpaper, 10 pt, conference]{ieeeconf}  %

\IEEEoverridecommandlockouts                              %

\pdfoutput=1

\usepackage{amssymb}
\usepackage{amsmath}
\usepackage{amsfonts}       %
\usepackage{booktabs}       %
\usepackage{balance}
\usepackage[utf8]{inputenc} %
\usepackage[T1]{fontenc}    %
\usepackage[pdftex, pdfstartview={FitV}, pdfpagelayout={TwoColumnLeft},bookmarksopen=true,plainpages = false, colorlinks=true, linkcolor=black, citecolor = black, urlcolor = black,filecolor=black , pagebackref=false,hypertexnames=false, plainpages=false, pdfpagelabels ]{hyperref}
\usepackage{url}            %
\usepackage{nicefrac}       %
\usepackage{microtype}      %
\usepackage[font=scriptsize]{caption}
\usepackage{subcaption}
\usepackage{graphicx}
\usepackage{epstopdf} %
\usepackage{mathtools}
\usepackage{marginnote}
\usepackage[dvipsnames]{xcolor}
\usepackage{float}
\usepackage[capitalize]{cleveref}
\usepackage[]{algorithmic}
\usepackage{makecell}
\usepackage{multirow}
\usepackage{paralist}
\floatstyle{boxed} \floatname{algorithm}{Algorithm}
\newfloat{algorithm}{t}{loa}[section]
\usepackage{xcolor}
\usepackage[nolist,nohyperlinks]{acronym}
\usepackage[sort,compress]{cite}

\usepackage{amsmath}
\usepackage{color}
\usepackage{soul}
\usepackage{xspace}
\usepackage[capitalize]{cleveref}
\usepackage{dsfont}
\usepackage{bm}
\usepackage{bbm}
\usepackage{nicefrac}  %
\usepackage{amsfonts}
\usepackage{amssymb}
\usepackage{array}
\usepackage{mathtools}
\usepackage{caption}
\usepackage{subcaption}
\usepackage{fontawesome}
\usepackage{thmtools}
\usepackage{thm-restate}
\usepackage{lipsum}
\usepackage{pifont}
\usepackage{makecell}
\usepackage{multirow}
\usepackage{afterpage}
\usepackage{tabularx} 
\usepackage{url}
\usepackage[normalem]{ulem}
\usepackage{wrapfig}  %
\usepackage{textcomp}

\Crefname{figure}{Figure}{Figures}
\crefname{figure}{Figure}{Figures}
\crefname{table}{Table}{Tables}
\crefformat{table}{Table~#1}
\crefformat{equation}{Equation (#1)}
\Crefformat{equation}{Equation (#1)}
\crefformat{algorithm}{Algorithm~#1}
\crefformat{appendix}{Appendix~#1}
\Crefformat{appendix}{Appendix~#1}
\crefformat{appendices}{Appendices~#1}
\Crefformat{appendices}{Appendices~#1}

\newcommand\sdots{\hbox to 1em{.\hss.\hss.}} %
\DeclareMathAlphabet\mathbfcal{OMS}{cmsy}{b}{n}  %

\newcommand{\xdes}{\mathbf{x}^\text{des}}    %
\newcommand{\xsafe}{\bar{\mathbf{x}}}   %
\newcommand{\usafe}{\bar{\mathbf{u}}}   %
\newcommand{\xest}{\hat{\mathbf{x}}}    %
\newcommand{\eest}{\boldsymbol{\xi}^{\text{est}}}  %
\newcommand{\ectrl}{\boldsymbol{\xi}^{\text{ctrl}}}%

\newif\ifcomments
\commentstrue

\ifcomments
	\newcommand{\aXX}[1]{\color{OliveGreen}AT: (#1)\color{black}\xspace}  %
	\newcommand{\XX}[1]{\color{red}JH: (#1)\color{black}\xspace}  %
\else
    \newcommand{\dXX}[1]{}  %
	\newcommand{\aXX}[1]{}  %
	\newcommand{\XX}[1]{}  %
\fi

\newcommand{\PreserveBackslash}[1]{\let\temp=\\#1\let\\=\temp}
\newcolumntype{C}[1]{>{\PreserveBackslash\centering}p{#1}}
\newcolumntype{R}[1]{>{\PreserveBackslash\raggedleft}p{#1}}
\newcolumntype{L}[1]{>{\PreserveBackslash\raggedright}p{#1}}
\crefformat{equation}{(#2#1#3)}

\title{\LARGE \bf
Output Feedback Tube MPC-Guided Data Augmentation for \\ Robust, Efficient Sensorimotor Policy Learning 
}

\author{Andrea Tagliabue, Jonathan P. How%
\thanks{All the authors are with the MIT Department of Aeronautics and Astronautics. \tt\{atagliab, jhow\}@mit.edu}%
}%

\begin{document}
\maketitle
\thispagestyle{empty}
\pagestyle{empty}
\begin{abstract}

\ac{IL} can generate computationally efficient \textit{sensorimotor} policies from demonstrations provided by computationally expensive model-based sensing and control algorithms. However, commonly employed \ac{IL} methods are often data-inefficient, requiring the collection of a large number of demonstrations and producing policies with limited robustness to uncertainties. In this work, we combine \ac{IL}  with an output feedback \ac{RTMPC} to co-generate demonstrations and a data augmentation strategy to \textit{efficiently} learn neural network-based \textit{sensorimotor} policies. Thanks to the augmented data, we reduce the computation time and the number of demonstrations needed by \ac{IL}, while providing robustness to sensing \textit{and} process uncertainty. We tailor our approach to the task of learning a trajectory tracking \textit{visuomotor} policy for an aerial robot, leveraging a 3D mesh of the environment as part of the data augmentation process. We numerically demonstrate that our method can  learn a robust \textit{visuomotor} policy from a single demonstration---a two-orders of magnitude improvement in demonstration efficiency compared to existing IL methods.

\end{abstract}
\section{INTRODUCTION}
\textit{Imitation learning} \cite{argall2009survey, ross2011reduction, pomerleau1989alvinn} is increasingly employed to generate computationally-efficient policies from computationally-expensive model-based sensing \cite{scaramuzza2011visual, zhang2014loam} and control \cite{tagliabue2021demonstration, loquercio2019deep, loquercio2021learning} algorithms for \textit{onboard} deployment. The key to this method is to leverage the inference speed of deep neural networks, which are trained to imitate a set of task-relevant expert demonstrations collected from the model-based algorithms. 
This approach has been used to generate efficient \textit{sensorimotor} policies  \cite{zhang2016learning, kahn2017plato, loquercio2019deep, kaufmann2020deep, ross2013learning} capable of producing control commands from raw sensory data, bypassing the computational cost of control \textit{and} state estimation, while providing benefits in terms of latency and robustness. Such sensorimotor policies have demonstrated impressive performance on a variety of tasks, including agile flight \cite{loquercio2019deep} and driving \cite{pan2020imitation}.

However, one of the fundamental limitations of existing \ac{IL} methods employed to produce sensorimotor policies (e.g., Behavior Cloning (BC) \cite{argall2009survey}, DAgger \cite{ross2011reduction}) is the overall number of demonstrations that must be collected from the model-based algorithm. This inefficiency hinders the possibility of generating policies by directly collecting data from the real robot, requiring the user to rely on accurate simulators. Furthermore, the need to repeatedly query a computationally expensive expert introduces high computational costs during training, requiring expensive training equipment. The fundamental cause of these inefficiencies is the need to achieve \textit{robustness} to sensing noise, disturbances, and modeling errors encountered during deployment, which can cause deviations of the policy's state distribution from its training distribution---an issue known as \textit{covariate shift}. 

\begin{figure}[t]
    \centering
    \includegraphics[width=0.8\columnwidth]{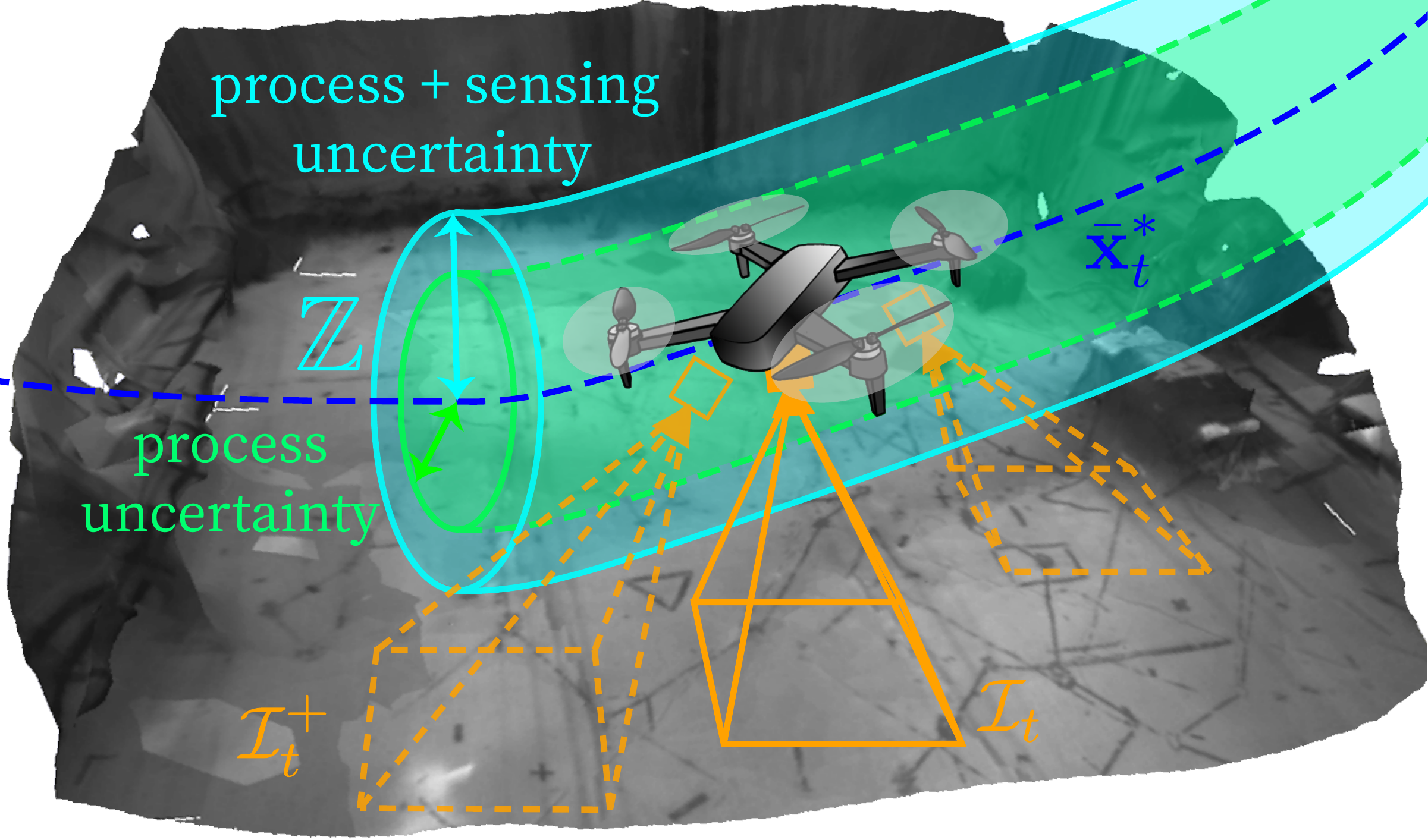}
    \caption{Proposed \textit{visuomotor} policy learning approach. We collect demonstrations from an output feedback robust tube MPC, which accounts for the effects of process \textit{and} sensing uncertainty via its tube section $\mathbb{Z}$. We use the tube to obtain a data augmentation strategy, employing a 3D mesh of the environment created from images $\mathcal{I}_t$ captured by an onboard camera during the demonstration collection phase. The augmented synthetic images $\mathcal{I}_t^+$ correspond to states sampled inside the tube, and the corresponding actions are obtained via the \textit{ancillary controller}, part of the tube MPC framework.}
    \label{fig:approach_overview}
    \vskip-4ex
\end{figure}

Strategies based on matching the disturbances between the training and deployment domains (e.g., 
\ac{DR} \cite{peng2018sim}) improve the robustness of the learned policy, but do not address the demonstration and computational efficiency challenges. Data augmentation approaches based on generating synthetic sensory measurements and corresponding stabilizing actions have shown promising results \cite{bojarski2016end, giusti2015machine}. However, they rely on handcrafted heuristics for the augmentation procedure and have been mainly studied on lower-dimensional tasks (e.g., steering a 2D Dubins car).

In this work, we propose a new framework that leverages a model-based robust controller, an output feedback \ac{RTMPC}, to provide \textit{robust} demonstrations \textit{and} a corresponding data augmentation strategy for \textit{efficient} \textit{sensorimotor} policy learning. Specifically, we extend our previous work where we show that \ac{RTMPC} can be leveraged to generate augmented data that enables efficient learning of a \textit{motor} control policy \cite{tagliabue2021demonstration}. This new work, summarized in \cref{fig:approach_overview}, employs an output feedback variant of \ac{RTMPC} and a model of the sensory observations to design a data augmentation strategy that also accounts for the uncertainty caused by \textit{sensing} imperfections during the demonstration collection phase. This approach enables efficient, robust \textit{sensorimotor} learning from a few demonstrations and relaxes our previous assumption \cite{tagliabue2021demonstration} that state information is available at training and deployment time. 

We tailor our method to the context of learning a \textit{visuomotor} policy, capable of robustly tracking given trajectories while using images from an onboard camera to control the position of an aerial robot. Our approach leverages a 3D mesh of the environment reconstructed from images to generate the observation model needed for the proposed augmentation strategy. This technique is additionally well-suited to be used with  photorealistic, data-driven simulation engines such as \textit{FlightMare} \cite{song2020flightmare} and \textit{FlightGoggles} \cite{guerra2019flightgoggles}, or SLAM and state estimation pipelines that directly produce 3D dense/mesh reconstructions of the environment, such as~\textit{Kimera}~\cite{rosinol2020kimera}.

\textbf{Contributions.} In summary, our work presents the following contributions:    
\begin{inparaenum}[a)]
        \item We introduce a new data augmentation strategy, an extension of our previous work \cite{tagliabue2021demonstration}, enabling efficient and robust learning of a \textit{sensorimotor} policy, capable to generate control actions directly from raw sensor measurements (e.g., images) instead of state estimates. Our approach is grounded in the output feedback \ac{RTMPC} framework theory, unlike previous methods that rely on handcrafted heuristics, and leverages a 3D mesh of the environment to generate augmented data.
        \item We demonstrate our methodology in the context of \textit{visuomotor} policy learning for an aerial robot, showing that it can track a trajectory from raw images with high robustness ($>90 \%$ success rate) after a \textbf{single demonstration}, despite sensory noise and disturbances. 
        \item We open-source our framework, available at \href{https://github.com/andretag}{https://github.com/andretag}.
\end{inparaenum}

\section{RELATED WORKS}
\textbf{Sensorimotor policy learning by imitating model-based experts.} 
Learning a sensorimotor policy by imitating the output of a model-based expert algorithm bypasses the computational cost of planning \cite{loquercio2021learning}, control \cite{tagliabue2021demonstration} and state estimation \cite{kahn2017plato}, with the potential of achieving increased robustness \cite{loquercio2019deep, dawson2022learning} and reduced latency with respect to conventional autonomy pipelines. The sensory input typically employed is based on raw images \cite{pomerleau1989alvinn, ross2013learning, pan2017agile} or pre-processed representations, such as feature tracks \cite{loquercio2019deep}, depth maps \cite{loquercio2021learning}, or intermediate layers of a CNN \cite{9560916}. 
Vision is often complemented with proprioceptive information, such as the one provided by IMUs \cite{loquercio2019deep} or wheel encoders \cite{pan2017agile}.
These approaches showcase the advantages of sensorimotor policy learning but do not leverage any data augmentation strategy. Consequently, they query the expert many times during the data collection phase, increasing time, number of demonstrations and computational effort to obtain the policy.

\textbf{Data augmentation for visuomotor and sensorimotor learning.}
Traditional data augmentation strategies for \textit{visuomotor} policy learning have focused on increasing a policy's generalization ability by applying perturbations \cite{hendrycks2019augmix} or noise \cite{antotsiou2021adversarial} directly in image space, without modifying the corresponding action. These methods do not directly address covariate shift issues caused by \textit{process} uncertainties.  The self-driving literature has developed a second class of visuomotor data augmentation strategies \cite{amini2020learning, bojarski2016end, giusti2015machine} capable of compensating covariate-shift issues. This class relies instead on first generating different views from synthetic cameras \cite{bojarski2016end, giusti2015machine} or data-driven simulators \cite{amini2020learning} and then computing a corresponding action via a handcrafted controller. These methods, however, rely on heuristics to establish relevant augmented views and the corresponding control action. Our work provides a more general methodology for data augmentation and demonstrates it on a quadrotor system (higher-dimensional than the planar self-driving car models).

\textbf{Output feedback RTMPC.} Model predictive control \cite{borrelli2017predictive} leverages a model of the system dynamics to generate actions that take into account state and actuation constraints. This is achieved by solving a constrained optimization problem along a predefined temporal horizon, using the model to predict the effects of future actions. Robust variants of MPC, such as \ac{RTMPC}, usually assume that the system is subject to additive, bounded \textit{process} uncertainty (e.g., disturbances, model errors). As a consequence, they modify nominal plans by either 
\begin{inparaenum}[a)]
\item assuming a worst-case disturbance~\cite{scokaert1998min, bemporad2003min}, or
\item employing an auxiliary (ancillary) controller. This controller maintains the system within some distance (``cross-section'' of a tube) from the nominal plan regardless of the realization of the disturbances \cite{mayne2005robust, lopez2019adaptive}.
\end{inparaenum}
\textit{Output feedback} \ac{RTMPC} \cite{mayne2006robust, kogel2017robust, lorenzetti2020simple} also accounts for the effects of \textit{sensing} uncertainty (e.g., sensing noise, imperfect state estimation).
Our work relies on an output feedback \ac{RTMPC} \cite{kogel2017robust, lorenzetti2020simple} to generate demonstrations and a data augmentation strategy. However, thanks to the proposed imitation learning strategy, our approach does not require solving the optimization problem online, reducing the onboard computational cost.

\section{PROBLEM STATEMENT}
Our goal is to generate a deep neural network sensorimotor policy $\pi_\theta$, with parameters $\theta$, to control a mobile robot (e.g., multirotor). The policy needs to be capable of tracking a desired reference trajectory given high-dimensional noisy sensor measurements, and has the form
\begin{equation}
\label{eq:policy}
    \mathbf u_t = \pi_\theta(\mathbf{o}_t, \mathbf{X}^\text{des}_t),
\end{equation}
where $t$ denotes the discrete time index, $\mathbf u_t$ represents the deterministic control actions, and $\mathbf{o}_t = (\mathcal{I}_t, \mathbf{o}_{\text{other},t})$ the high-dimensional, noisy sensor measurements, comprised of an image $\mathcal{I}_t$ captured by an onboard camera, and other noisy measurements $\mathbf{o}_\text{other,t}$ (e.g., attitude, velocity). The $N+1$ steps  of the reference trajectory are written as $\mathbf{X}^\text{des}_t = \{\xdes_{0|t},\dots,\xdes_{N|t}\}$, where $\xdes_{i|t}$ indicates the desired state at the future time $t+i$, as given at the current time $t$, and $N > 0$ represents the total number of given future desired states.
Our objective is to \textit{efficiently} learn the policy parameters $\hat{\theta}^*$ by leveraging \ac{IL} and demonstrations provided by a model-based controller (\textit{expert}). 

\noindent
\textbf{System model.}
We assume available a model of the dynamics of the robot, described by a set of linear (e.g., via linearization), discrete and time-invariant equations: 
\begin{equation}
    \mathbf x_{t+1} = \mathbf A \mathbf x_{t} + \mathbf B  \mathbf u_{t} + \mathbf w_{t} \\
\label{eq:linearized_dynamics}
\end{equation}
where matrices $\mathbf A \in \mathfrak{R}^{n_x \times n_x}$ and $\mathbf B \in \mathfrak{R}^{n_x \times n_u}$ represent the system dynamics, 
$\mathbf x_t  \in \mathbb{X} \subset \mathfrak{R}^{n_x}$ represents the state of the system, and $\mathbf u_t \in \mathbb{U} \subset \mathfrak{R}^{n_u}$ represents the control inputs. The system is subject to state and inputs constraint $\mathbb{X}$ and $\mathbb{U}$, assumed to be convex polytopes containing the origin \cite{mayne2006robust}. The quantity $\mathbf w_t \in \mathbb{W} \subset \mathfrak{R}^{n_x}$ in \eqref{eq:linearized_dynamics} captures time-varying additive \textit{process} uncertainties. This includes disturbances and model errors that the system may encounter at deployment time, or the ones caused by the \textit{sim2real} gap when collecting demonstrations using a simulator. It can additionally capture disturbances encountered at deployment time that are not present when collecting demonstrations from a real robot in a controlled environment (e.g., \textit{lab2real} gap). These uncertainties are assumed to belong to a bounded set $\mathbb{W}$, assumed to be a polytope containing the origin \cite{mayne2006robust}.

\noindent
\textbf{Sensing model.} We assume that the model describing the measurement $\bar{\mathbf{o}}_t \in \mathfrak{R}^{n_o}$ used by the expert is:
\begin{equation}
\label{eq:obs_model}
    \bar{\mathbf{o}}_t = 
    \begin{bmatrix}
        \mathbf o_{\text{pos}, t} \\
        \mathbf o_{\text{other}, t} \\
    \end{bmatrix}
    = \mathbf C
    \mathbf x_t 
    + \mathbf v_t,
\end{equation}
where $\mathbf C \in \mathfrak{R} ^{n_x \times n_o}$, and  $\mathbf v_t = [\mathbf v_{\text{\text{cam}},t}^T, \mathbf v_{\text{other},t}^T]^T \in \mathbb{V} \subset \mathfrak{R}^{n_o}$ represents additive \textit{sensing} uncertainties (e.g., noise, biases), sampled from a bounded set $\mathbb{V}$. The expert has access to a vision-based position estimator $g_{\text{cam}}$ that produces a noisy estimate $\mathbf o_\text{pos} \in \mathfrak{R}^3$ of the position $\boldsymbol{p}_t \in \mathfrak{R}^3$ of the robot from an image:
\begin{equation}
\label{eq:position_estimator}
    \mathbf o_{\text{pos}, t} =  g_{\text{cam}}(\mathcal{I}_t)  = \boldsymbol{p}_t + \mathbf v_{\text{\text{cam}},t}, 
\end{equation}
where $\mathbf{v}_{\text{cam},t}$ is the associated noise/uncertainty. $\mathbb{V}$ can be obtained via system identification, and/or prior knowledge on the accuracy of $g_\text{cam}$.

\noindent
\textbf{State estimator model.}
\label{sec:state_observer}
We assume that the dynamics of the state estimator employed by the expert during the demonstration collection phase can be approximated by the linear (Luenberger) observer
\begin{equation} \label{eq:estimator}
    \xest_{t+1} = \mathbf A \xest_t + \mathbf B \mathbf u_t + \mathbf L ( \bar{\mathbf{o}}_t - \hat{\mathbf o}_t), \quad \hat{\mathbf o}_t = \mathbf C \xest_t,
\end{equation}
where $\hat{\mathbf{o}}_t$ is the predicted observation, and $\xest_t \in \mathfrak{R}^{n_x}$ is the estimated state. $\mathbf L \in \mathbb  R^{n_x \times n_o}$ is the observer gain matrix, and is chosen so that the matrix $\mathbf A - \mathbf L \mathbf C$ is Schur stable. The observability index of the system $(\mathbf A, \mathbf C)$ is assumed to be $1$, meaning that full state information can be retrieved from a single noisy measurement. In this case, the observer plays the critical role of filtering out the effects of noise and sensing uncertainties. Finally, we assume that the state estimation dynamics and noise sensitivity of the learned policy will approximately match the ones of the observer.

\section{METHODOLOGY}
The overall idea of our method is to generate trajectory tracking demonstrations from noisy sensor measurements using an output feedback \ac{RTMPC} expert combined with a state estimator \cref{eq:estimator}, leveraging the defined process and sensing models.
The chosen output feedback \ac{RTMPC} framework is based on \cite{lorenzetti2020simple}, with its cost function modified to track a desired trajectory, and it is summarized in \cref{sec:tube_mpc}. This framework provides:
\begin{inparaenum}[a)]
    \item \textit{robust} demonstrations, taking into account the effect of process \textit{and} sensing uncertainties, and
    \item a corresponding data augmentation strategy, key to overcome efficiency and robustness challenges in \ac{IL}.
\end{inparaenum}
Demonstrations can be collected via \ac{IL} methods such as BC \cite{pomerleau1989alvinn} or DAgger \cite{ross2011reduction}.
Combined with the augmented data provided by our strategy, the collected demonstrations are then used to train a policy \cref{eq:policy} via supervised regression. \cref{sec:evaluation} will tailor our methodology to control a multirotor system.

\subsection{Mathematical preliminaries}
\noindent
\textbf{Set operations.} Given the convex polytopes $\mathbb{A} \subset \mathfrak{R}^{n}, \mathbb{B} \subset \mathfrak{R}^{n}$ and the linear mapping $\mathbf{C} \in \mathfrak{R}^{m \times n}$, we define:
\begin{enumerate}[a)]
\item Minkowski sum: $\mathbb{A} \oplus \mathbb{B} \coloneqq \{\mathbf a + \mathbf b \in \mathfrak{R}^n \:|\: \mathbf a \in \mathbb{A}, \: \mathbf b \in \mathbb{B} \}$
\item Pontryagin diff.: $\mathbb{A} \ominus \mathbb{B} \coloneqq \{\mathbf c \in \mathfrak{R}^n \:|\: \mathbf{c + b} \in \mathbb{A}, \forall \mathbf  b \in \mathbb{B} \}$ 
\end{enumerate}
\noindent
\textbf{Robust Positive Invariant (RPI) set.} Given a closed-loop Schur-stable system $\mathbf A \in \mathfrak{R}^{n \times n}$ subject to the uncertainty $\mathbf{d}_t \in \mathbb{D} \subset \mathfrak{R}^n$: 
$
    \boldsymbol{\xi}_{t+1} = \mathbf A \boldsymbol{\xi}_t + \mathbf{d}_t,
$
we define the \textit{minimal} Robust Positive Invariant (RPI) set $\mathbb{S}$ as the \textit{smallest} set satisfying the following property:
\begin{equation}
    \boldsymbol{\xi}_0 \in \mathbb{S} \implies \boldsymbol{\xi}_t \in \mathbb{S}, \; \forall \; \mathbf d_t \in \mathbb{D}, \; t > 0.
\end{equation}
This means that if the initial state $\boldsymbol{\xi}_0$ of the system starts in the RPI set $\mathbb{S}$, then it will never escape from such set, regardless of the realizations $\mathbf d_t$ of the disturbances in $\mathbb{D}$.

\subsection{Output feedback robust tube MPC for trajectory tracking}
\label{sec:tube_mpc}
The objective of the output feedback \ac{RTMPC} for trajectory tracking is to regulate the system in \cref{eq:linearized_dynamics} along a given \textit{desired} trajectory $\mathbf{X}^\text{des}_t$, while ensuring satisfaction of the state and actuation constraints $\mathbb X, \mathbb U$ regardless of the process noise $\mathbf w$ and sensing noise $\mathbf v$ in \cref{eq:linearized_dynamics} and \cref{eq:obs_model}, respectively. 

\noindent
\textbf{Optimization problem.} To fulfill these requirements, the RTMPC computes at every timestep $t$ a sequence of ``safe'' reference states $\mathbf{\bar{X}}_t = \{\xsafe_{0|t},\dots,\xsafe_{N|t}\}$ and actions $\mathbf{\bar{U}}_t = \{\usafe_{0|t},\dots,\usafe_{N-1|t}\}$ along a planning horizon of length $N+1$. This is achieved by solving the optimization problem:
\begin{equation} \label{eq:ofrtmpc}
\begin{split}
    \mathbf{\bar{U}}_t^*, \mathbf{\bar{X}}_t^*
    = \underset{\mathbf{\bar{U}}_t, \mathbf{\bar{X}}_t}{\text{argmin}} & 
        \| \mathbf e_{N|t} \|^2_\mathbf{P} + 
        \sum_{i=0}^{N-1} 
            \| \mathbf e_{i|t} \|^2_\mathbf{Q} + 
            \| \mathbf u_{i|t} \|^2_\mathbf{R} \\
    \text{subject to} \:\: & \xsafe_{i+1|t} = \mathbf A \xsafe_{i|t} + \mathbf B \usafe_{i|t},  \\
    & \xsafe_{i|t} \in \bar{\mathbb{X}}, \:\: \usafe_{i|t} \in \bar{\mathbb{U}}, \\
    & \mathbf{\bar{x}}_{N|t} \in \mathbb{X}_N, \:\: \xest_t \in \mathbb{Z} \oplus \xsafe_{0|t},  \\
\end{split}
\end{equation}
where $\mathbf e_{i|t} = \xsafe_{i|t} - \xdes_{i|t}$ represents the trajectory tracking error. The positive definite matrices $\mathbf{Q}$ (size $n_x \times n_x$) and $\mathbf{R}$ (size $n_u \times n_u$) are tuning parameters, and define the trade-off between deviations from the desired trajectory and actuation usage. A terminal cost $\| \mathbf e_{N|t} \|^2_\mathbf{P}$ and terminal state constraint $\mathbf{\bar{x}}_{N|t} \in \bar{\mathbb{X}}_N$ are introduced to ensure stability \cite{lorenzetti2020simple}, where $\mathbf P$ (size $n_x \times n_x$) is a positive definite matrix, obtained by solving the infinite horizon optimal control LQR problem, using the system dynamics $\mathbf A$, $\mathbf B$ in \cref{eq:linearized_dynamics} and the weights $\mathbf Q$ and $\mathbf R$. The ``safe'' references $\mathbf{\bar{X}}_t$, $\mathbf{\bar{U}}_t$ are generated based on the nominal (e.g., not perturbed) dynamics, via the constraints $\xsafe_{i+1|t} = \mathbf A \xsafe_{i|t} + \mathbf{B} \usafe_{i|t}$.

\noindent
\textbf{Ancillary controller.} The control input for the real system is obtained via a feedback controller, called \textit{ancillary controller}:
\begin{equation}
\label{eq:ancillary_controller}
    \mathbf u_t = \mathbf \usafe^*_{t} + \mathbf K (\xest_t - \xsafe^*_{t}),
\end{equation}
where $\usafe^*_{0|t} = \usafe^*_t$ and $\xsafe^*_{0|t} = \xsafe^*_t$. The matrix $\mathbf{K}$ is computed by solving the LQR problem with $\mathbf A$, $\mathbf B$, $\mathbf Q$, $\mathbf{R}$. This controller ensures that the system remains inside a set $\mathbb{Z}$ (``cross-section'' of a \textit{tube}) centered around $\xsafe_t^*$ regardless of the realization of process and sensing uncertainty, provided that the state of the system starts in such tube (constraint $\xest_t \in \mathbb{Z} \oplus \xsafe_{0|t}$).

\noindent
\textbf{Tube and robust constraint satisfaction.}
Process and sensing uncertainties are taken into account by tightening the constraints $\mathbb{X}$, $\mathbb{U}$ by an amount which depends on $\mathbb{Z}$, obtaining $\bar{\mathbb{X}}$ and $\bar{\mathbb{U}}$. 

We start by describing how to compute the set $\mathbb{Z}$, which depends on the errors caused by uncertainties  when employing the ancillary controller \cref{eq:ancillary_controller} and the state estimator \cref{eq:estimator}. The system considered in \cref{eq:linearized_dynamics} and \cref{eq:obs_model} is subject to \textit{sensing} $\mathbb{V}$  and \textit{process} $\mathbb{W}$ uncertainties, sources of two errors: 
\begin{inparaenum}[a)]
\item the state estimation error $\eest_t = \mathbf x_t - \xest_t$, which captures the deviations of the state estimate $\xest_t$ from the true state $\mathbf x_t$ of the system;
\item the control error $\ectrl_t = \xest_t - \xsafe_t^*$, which captures instead deviations of  $\xest_t$ from the reference plan $\xsafe_t^* = \xsafe^*_{0|t}$. 
\end{inparaenum}
These errors can be combined in a vector $\boldsymbol{\xi}_t = [{\eest_t}^T, {\ectrl_t}^T ]^T$ whose dynamics evolve according to (see \cite{kogel2017robust, lorenzetti2020simple}):
\begin{equation} 
\label{eq:error}
\boldsymbol{\xi}_{t+1}  = \mathbf A_\xi \boldsymbol{\xi}_t  + \boldsymbol{\delta}_t, \; \boldsymbol{\delta}_t \in \mathbb{D}\\
\end{equation}
\begin{equation*}
\mathbf A_\xi = \begin{bmatrix}
\mathbf A- \mathbf L \mathbf C & \mathbf{0}_{n_x} \\ \mathbf L \mathbf C & \mathbf A + \mathbf B \mathbf K
\end{bmatrix}, \quad 
\mathbb{D} = \begin{bmatrix}
\mathbf I_{n_x} & -\mathbf L \\ \mathbf{0}_{n_x} & \mathbf L 
\end{bmatrix}\begin{bmatrix}
\mathbb{W} \\ \mathbb{V}
\end{bmatrix}.
\label{eq:autonomous_system}
\end{equation*}
We observe that, by design, $\mathbf A_\xi$ is Schur-stable, and $\mathbb{D}$ is a convex polytope. Then, the possible set of state estimation and control errors correspond to the minimal RPI set $\mathbb{S}$ associated with the system $\mathbf{A}_\xi$ under the uncertainty set $\mathbb{D}$. 

We now introduce the error between the true state $\mathbf x_t$ and the reference state $\xsafe_t^*$:
$
\boldsymbol{\xi}^{\text{tot}} = \mathbf{x}_t - \xsafe_t^* = \boldsymbol{\xi}^{\text{ctrl}} + \boldsymbol{\xi}^{\text{est}}.
$
As a consequence, the set (\textit{tube}) $\mathbb{Z}$ that describes the possible deviations of the true state $\mathbf{x}_t$ from the reference $\xsafe_t^*$ is:
\begin{equation}
\label{eq:tube}
    \mathbb{Z} = 
    \begin{bmatrix}
    \boldsymbol{I}_{n_x} & \boldsymbol{I}_{n_x} \\
    \end{bmatrix}\mathbb{S}.
\end{equation}
Finally, the effects of noise and uncertainties can be taken into account by tightening the constraints of an amount: 
\begin{equation}
\bar{\mathbb{X}} \coloneqq \mathbb{X}\ominus
\mathbb{Z}, \quad \bar{\mathbb{U}} \coloneqq \mathbb{U} \ominus\begin{bmatrix}
\boldsymbol{0}_{n_x} & \mathbf K
\end{bmatrix} \mathbb{S}.
\end{equation}

\subsection{Tube-guided data augmentation for visuomotor learning}
\label{sec:data_augmentation}
\noindent
\textbf{Imitation learning objective.} 
We assume that the given output feedback RTMPC controller in \cref{eq:ofrtmpc},  \cref{eq:ancillary_controller} and the state observer in \cref{eq:obs_model} can be represented via a policy $\pi_{\theta^*}$ \cite{laskey2017dart}, where $\pi$ is of the form in \cref{eq:policy}, with parameters $\theta^*$. Our objective is to design an \ac{IL} and data augmentation strategy to efficiently learn the parameters $\hat{\theta}^*$ for the sensorimotor policy \cref{eq:policy}, collecting demonstrations from the expert. 
Using \ac{IL}, this objective is achieved by minimizing the expected value of a distance metric $J(\theta, \theta^*|\tau)$ over a distribution of trajectories $\tau$:
\begin{equation}
    \hat{\theta}^* = \text{arg}\min_{\theta} \mathbb{E}_{p(\tau|\pi_\theta, \mathcal{K}_\text{target})}
    [J(\theta, \theta^*|\tau)],
    \label{eq:il_obj_target}
\end{equation}
where $J(\theta, \theta^*|\tau)$ captures the differences between the trajectories generated by the expert $\pi_{\theta^*}$ and the trajectories produced by the learner $\pi_\theta$. 
The variable $\tau := ((\mathbf o_0, \mathbf u_0, \mathbf X^\text{des}_0), \dots, (\mathbf o_T, \mathbf u_T, \mathbf X^\text{des}_T))$ denotes a $T+1$ step (observation, action, reference) trajectory sampled from the distribution $p(\tau|\pi_\theta, \mathcal{K}_\text{target})$. 
Such distribution represents all the possible trajectories induced by the learner policy $\pi_{\theta}$ in the deployment environment $\mathcal{K}_\text{target}$, under all the possible instances of the process \textit{and} sensing uncertainties. 
In this work, the chosen distance metric is the MSE loss:
\begin{equation}
    J(\theta, \theta^*|\tau) = \frac{1}{T}\sum_{t=0}^{T-1}\| \pi_\theta(\mathbf o_t, \mathbf{X}_t^\text{des}) - \pi_{\theta^*}(\mathbf o_t, \mathbf{X}_t^\text{des})\|_2^2.
    \label{eq:mse_loss}
\end{equation}
As observed in \cite{tagliabue2021demonstration}, the presence of uncertainties in the deployment domain $\mathcal{K}_\text{target}$ makes the setup challenging, as demonstrations are usually collected in a training environment $\mathcal{K}_\text{source}$ that presents only a subset of all the possible uncertainty realizations. At deployment time, uncertainties may drive the learned policy towards the region of the input space for which no demonstrations have been provided (covariate shift), with potentially catastrophic consequences. 

\noindent
\textbf{Tube and ancillary controller for data augmentation.}
To overcome these limitations, we design a data augmentation strategy that takes into account and compensates for the effects of \textit{process and sensing uncertainties} encountered in $\mathcal{K}_\text{target}$, and that it is capable of producing the augmented input-output data needed to learn the sensorimotor policy \cref{eq:policy}. 
We do so by extending our previous approach \cite{tagliabue2021demonstration}, named \ac{SA}, which provides a strategy to efficiently learn a policy robust to \textit{process} uncertainty. \ac{SA} relies on recognizing that the tube $\mathbb{Z}$, as computed via a \ac{RTMPC}, represents a model of the states that the system may visit when subject to process uncertainties. The ancillary controller \cref{eq:ancillary_controller} provides a computationally efficient way to generate the actions to take inside the tube, as it guarantees that the system remains inside the tube for every possible realization of the process uncertainty. 

Our new approach, named \ac{VSA}, employs the output feedback variant of \ac{RTMPC} in \cref{sec:tube_mpc}, as its tube represents a model of the states that might be visited by the system when subject to process \textit{and sensing} uncertainty, and its ancillary controller provides a way to efficiently compute corresponding actions, guaranteeing that the system remains inside the tube $\mathbb{Z} \oplus \xsafe^*_t$ in \cref{eq:tube} for every possible realization of $\mathbf{w}_t \in \mathbb{W}$ \textit{and} $\mathbf{v}_t \in \mathbb{V}$. These properties can be leveraged to design a data augmentation strategy while collecting demonstrations in the source domain $\mathcal{K}_\text{source}$.

More specifically, let $\tau^\text{aux}$ be a $T+1$ step demonstration generated by the output feedback RTMPC in \cref{sec:tube_mpc}, collected in the training domain $\mathcal{K}_\text{source}$: 
\begin{multline}
    \tau^\text{aux} = ((\mathbf o_0, \mathbf u_0, (\usafe_0^*, \xsafe_0^*), \mathbf{X}^\text{des}_0), \dots \\
    \dots, (\mathbf o_T, \mathbf u_T, (\usafe_T^*, \xsafe_T^*), \mathbf{X}^\text{des}_T)).
\end{multline}
Then, at every timestep $t$ in the collected demonstration $\tau^\text{aux}$, we can compute extra (state, action) pairs $(\mathbf{x}^+, \mathbf{u}^+)$ by sampling extra states from inside the tube $\mathbf x^+ \in \xsafe_t^* \oplus \mathbb{Z}$, and the corresponding robust control action $\mathbf u^+$:
\begin{equation}
\mathbf u^+ = \mathbf \usafe_t^* +  \mathbf K (\mathbf{x}^+ - \xsafe_t^*).
\label{eq:tubempc_feedback_policy}
\end{equation}
 
\noindent
\textbf{Augmented sensory data generation.}
The output feedback RTMPC-based augmentation strategy described so far provides a way to generate extra (state, action) samples for data augmentation, but not the sensor observations $\mathbf{o}^+ = (\mathcal{I}^+, \mathbf{o}_{\text{other}}^+)$ required as input of the sensorimotor policy \cref{eq:policy}. We overcome this issue by employing observation models \cref{eq:obs_model} available for the expert, obtaining a way to generate  extra observations $\mathbf{o}^+$, given extra states $\mathbf{x}^+$. 

In the context of learning the visuomotor policy \cref{eq:policy}, we generate  synthetic camera images $\mathcal{I}^+$ via an inverse pose estimator $g^{-1}_{\text{\text{cam}}}$, mapping camera poses $\mathbf T_{IC}$ to image observations $\mathcal{I}$
\begin{equation}
    \mathcal{I}^+ = g_{\text{\text{cam}}}^{-1}(\mathbf T_{IC}^+)
\end{equation}
where $\mathbf T_{IC}$ denotes a homogeneous transformation matrix \cite{barfoot2017state} from a world (inertial) frame $I$ to a camera frame $C$.
The inverse camera model $g^{-1}_{\text{\text{cam}}}$ can be obtained by generating a 3D reconstruction of the environment from the sequence of observations obtained in the collected demonstrations, and by modeling the intrinsic and extrinsic parameters of the onboard camera used by the robot \cite{barfoot2017state}. Extra camera poses $\textbf T_{IC}^+$ can be obtained from the extra state vector $\mathbf{x}^+$, assuming that the state contains position and orientation information
\begin{equation}
    \textbf T_{IC}^+ = \textbf T_{IB}(\mathbf x^+) \textbf T_{BC},
\end{equation}
where $\mathbf T_{IB}$ represents the homogeneous transformation expressing the pose of a robot-fixed body frame $B$ with respect to $I$. $\mathbf T_{BC}$ describes a known rigid transformation between the body of the robot $B$ and the optical frame of the camera $C$.
The full extra observation $\mathbf{o}^+$ can be obtained by additionally computing $\mathbf{o}_{\text{other}}^+$ via the measurement model \cref{eq:obs_model}, assuming zero noise: 
\begin{equation}
    \mathbf {o}^+ = (\mathcal{I}^+, \mathbf{o}_{\text{other}}^+), \; \; \mathbf{o}_{\text{other}}^+ = \mathbf{S} \mathbf{C} \mathbf{x}^+,
\end{equation}
where $\mathbf{S}$ is a selection matrix. 

\section{EVALUATION}
\subsection{Visuomotor trajectory tracking on a multirotor}
\label{sec:evaluation}
\noindent
\textbf{Task setup.} 
We consider the task of learning a visuomotor trajectory tracking policy for a multirotor, starting from slightly different initial states. The chosen trajectory is an eight-shape (lemniscate), with a velocity of up to $3.5$ m/s and a total duration of $30.0$ s. We assume that the robot is equipped with a downward-facing monocular camera, and it has available onboard noisy velocity $_I \boldsymbol{v} \in \mathfrak{R}^3$ and roll $_I \varphi$, pitch $_I \vartheta$ information, corrupted by measurement noise $\mathbf v_{\text{other},t}$:
\begin{equation}
   \mathbf o_{\text{other}, t} =  [_I \boldsymbol{v}^T, _I \varphi, _I \vartheta]^T + \mathbf v_{\text{other},t}.
\end{equation}
$I$ denotes a yaw-fixed, gravity aligned inertial reference frame in which the quantities are expressed, and $\mathbf v_\text{other}$ is a zero-mean, Gaussian noise with three standard deviations $3\boldsymbol{\sigma}_\text{other} = [0.2, 0.2, 0.2, 0.05, 0.05]^T$ (units in $m/s$ for velocity, and $rad$ for tilt). Information on the position $_I \boldsymbol{p} \in \mathfrak{R}^3$ of the robot is not provided to the learned policy, and it must be inferred from the onboard camera. The output feedback \ac{RTMPC} has instead additionally access to noisy position information corrupted by additive, zero mean Gaussian noise $\mathbf{v}_\text{cam}$ with three standard deviations $3\boldsymbol{\sigma}_{\text{cam}} = [0.2, 0.2, 0.4]^T$ ($m$). 
Demonstrations are collected from the output feedback RTMPC expert in a nominal domain $\mathcal{K}_\text{source}$ that presents sensing noise and model uncertainties (e.g., due to drag, unmodeled attitude dynamics). We consider two deployment environments $\mathcal{K}_\text{target}$, one with sensing noise, matching $\mathcal{K}_\text{source}$, and one that additionally presents wind-like force disturbances $\mathbf{f}_\text{wind}$, with magnitude corresponding to $15\%$-$20\%$ of the weight of the multirotor ($12.75$ N). During training and deployment, the robot should not collide with the walls/ceiling nor violate additional safety limits (maximum velocity, roll, pitch).

\noindent
\textbf{Output feedback \ac{RTMPC} design.}
To design the output feedback RTMPC, we employ a hover-linearized model of a multirotor\cite{kamel2017linear}. The state of the robot ($n_x = 8$) is:
\begin{equation}
    \mathbf x = [ _I\boldsymbol p^T, _I \boldsymbol v^T, _I \varphi, _I \vartheta ]^T,
\end{equation}
while the output ($n_u =3$) $\mathbf{u}_t$ corresponds to roll, pitch and thrust commands, which a cascaded attitude controller executes. Safety limits and position limits are encoded as state constraints. The sampling period for the controller and prediction model is set to $T_s = 0.1$ s, and the prediction horizon is $N = 30$, ($3.0$ s).
The tube $\mathbb{Z}$ is computed assuming the observation model in \cref{eq:obs_model} to be:
$
    \bar{\mathbf{o}}_t = \mathbf x_t + \mathbf v_t,
$
where the bounded measurement uncertainty matches the three standard deviations of the noise encountered in the environment: $\mathbb{V} = \{ \mathbf{v} \; | \; \|\mathbf v\|_\infty \leq 3 [ \boldsymbol{\sigma}_{\text{cam}}^T, \boldsymbol{\sigma}_\text{other}^T]^T \}$. The observer gain matrix $\mathbf L$ is computed by assuming fast state estimation dynamics, by placing the poles of the estimation error dynamics $\mathbf{A - LC}$ at $30.0$ rad/s.
\begin{figure}
    \centering
    \includegraphics[width=\columnwidth]{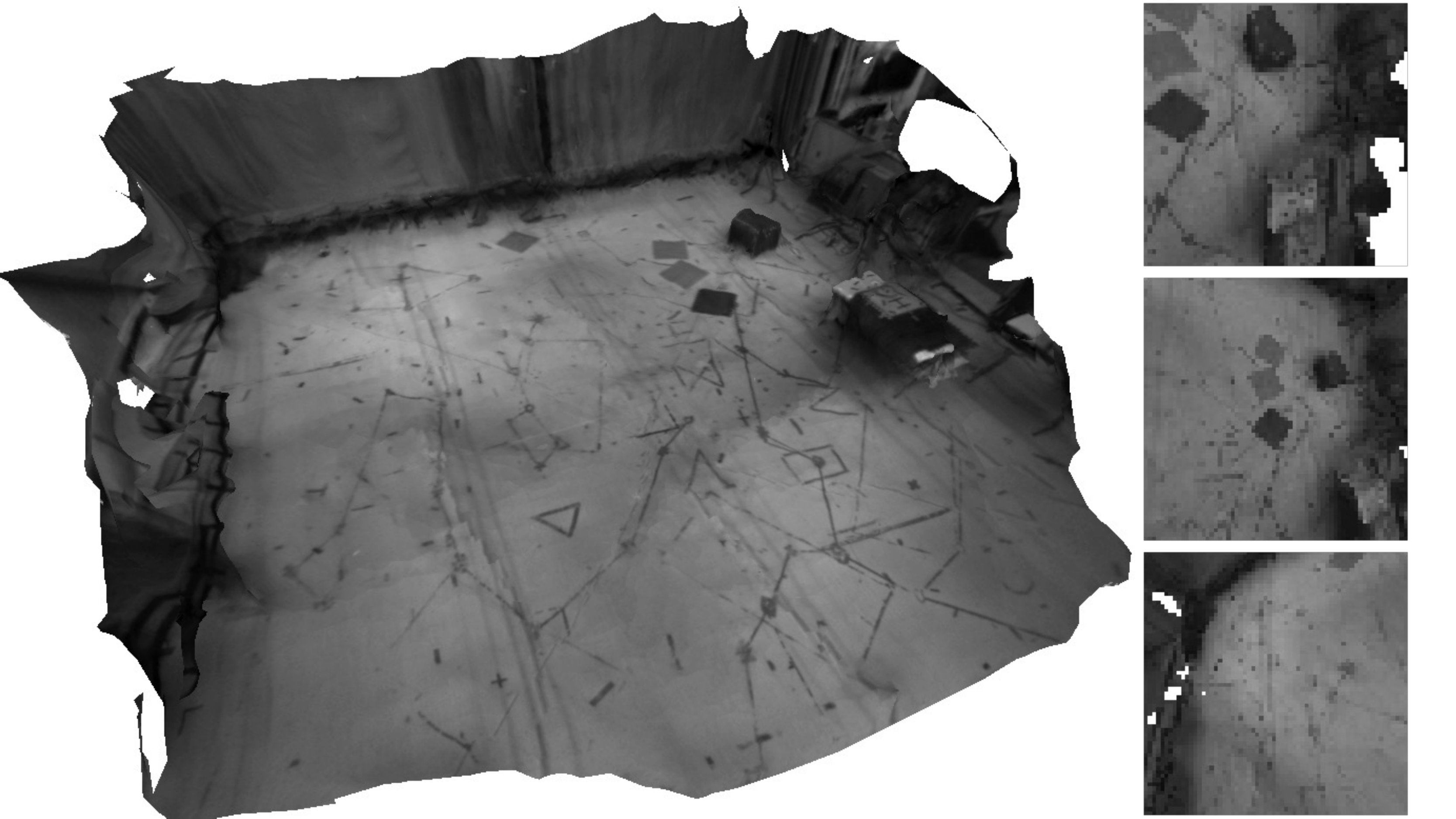}
    \caption{\textit{Left:} 3D reconstruction of the training environment employed to generate data augmentation. The camera images used to generate the reconstruction are obtained from a single demonstration collected with the real robot. \textit{Right:} Images generated by the simulated onboard camera along the considered trajectory in the 3D mesh of the environment. If the camera is pointing at regions where the reconstruction is not available (e.g., top and bottom right images), the area of missing data is left \textit{featureless} (white), without providing any additional information useful for position estimation.}
    \label{fig:3d_environment}
    \vskip-4ex
\end{figure}
The process uncertainty $\mathbb{W}$ is set to correspond to a bounded force-like disturbance of magnitude equal to $25\%$ of the weight of the multirotor, capturing the physical limits of the platform. Using these priors on uncertainties, sensing noise and the system \cref{eq:error}, we compute an approximation of the tube $\mathbb{Z}$ via Monte Carlo simulations, randomly sampling instances of the uncertainties. The RPI set $\mathbb S$ is obtained by computing an axis-aligned bounding box approximation of the error trajectories.

\begin{figure}[t]
    \centering
    \includegraphics[trim={110 602 190 52},clip, width=\columnwidth]{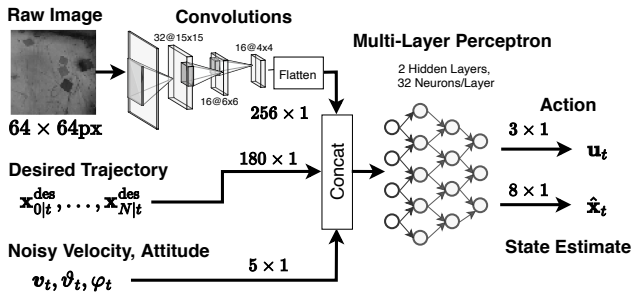}
    \caption{Architecture of the employed visuomotor policy. The policy takes as input a raw camera image, a desired trajectory $\xdes_{0|t}, \dots, \xdes_{N|t}$ and noisy measurements of the velocity $\boldsymbol{v}_t$ and tilt (roll $\varphi_t$, pitch $\vartheta_t$) of the multirotor. It outputs an action $\textbf{u}_t$, representing a desired roll, pitch and thrust set-points for the cascaded attitude controller. The policy additionally outputs an estimate of the state $\xest_t$, which was found useful to promote learning of features relevant for position estimation.}
    \label{fig:policy_architecture}
    \vskip-4ex
\end{figure}

\noindent
\textbf{Policy architecture.}
The architecture of the learned policy \cref{eq:policy} is represented in \cref{fig:policy_architecture}. The policy takes as input the raw image produced by the onboard camera, the desired trajectory, and noisy attitude and velocity information $\mathbf{o}_\text{other}$. It employs three convolutional layers, which map the raw images into a lower-dimensional feature space; these features are then combined with the additional velocity and attitude inputs, and passed through a series of fully connected layers. The generated action $\mathbf u_t$ is then clipped such that $\mathbf u_t \in \mathbb{U}$. To promote learning of internal features relevant to estimating the robot's state, the output of the policy is augmented to predict the current state $\xest$ (or $\mathbf{x}^+$ for the augmented data), modifying \cref{eq:mse_loss} accordingly. This output is not used at deployment time, but it was found to improve the performance. 

\subsection{Dense reconstruction of the environment}
We generate a 3D mesh of the environment needed for data augmentation via \ac{VSA} from a sequence of real images collected by the onboard downward-facing camera of the Qualcomm Snapdragon Flight Pro \cite{Qualcomm2022Snapdragon} board mounted on a multirotor. The images are obtained by executing the trajectory that we intend to learn, which helps us validate the assumption that it is possible to collect a quantity and quality of images sufficient to create an accurate 3D mesh from a single demonstration.
Overall we employ $145$ black and white images of size $640 \times 480$ pixels. The 3D reconstruction of the considered indoor flight space and the resulting mesh, shown in \cref{fig:3d_environment} (\textit{left}), are generated via the 3D photogrammetry pipeline provided by \textit{Meshroom} \cite{griwodz2021alicevision}. 
The scale of the generated reconstruction, which is unobservable from the monocular images, is set manually via prior knowledge of the dimensions of the environment. 
The obtained mesh is then integrated with our multirotor simulator. We additionally simulate a pinhole camera, whose extrinsic and intrinsic are set to approximately match the parameters of the real onboard camera. \cref{fig:3d_environment} (\textit{right}) shows examples of synthetic images from our simulator.

\subsection{Choice of baselines, training setup and hyperparameters}
Demonstration collection, training and evaluation is performed in simulation, employing a realistic full nonlinear multirotor model (e.g., \cite{kamel2017linear}).
We apply our data augmentation strategy, \ac{VSA}, to BC and DAgger, comparing their performance without any data augmentation. We additionally combine BC and DAgger with Domain Randomization (DR); in this case, during demonstration collection, we apply wind disturbances with magnitude sampled from $10\%$-$110\%$ of the maximum wind force at deployment.
We set $\beta$, hyperparameter of DAgger controlling the probability of using actions from the expert instead of the learner policy, to be $\beta=1$ for the first set of demonstrations, and $\beta=0$ otherwise. 
For every method, we perform our analysis in $M$ iterations, where at every $m$th iteration we:
\begin{inparaenum}[(i)]
    \item collect $K$ new demonstrations via the output feedback RTMPC expert and the state estimator;
    \item train a policy with randomly initialized weights using all the demonstrations collected so far, and 
    \item evaluate the obtained policy for $20$ times, under different realizations of the uncertainty and initial states of the robot.
\end{inparaenum}
We record the time taken to perform phase (i)-(ii) at the $m$th iteration, denoted as $T_m^\text{iter}$.
The entire analysis is  repeated across $10$ random seeds. We use $M=2$, $K=1$ for methods using \ac{VSA}, while $M=15$, $K=10$ for all the other methods, in order to speed up their computation. When using \ac{VSA}, we additionally train the policy incrementally, i.e., we only use the newly collected demonstrations and the corresponding augmented data to update the previously trained policy. All the policies are trained for $50$ epochs using the ADAM optimizer, with a learning rate of $0.001$ and a batch size of  $32$. When using \ac{VSA}, we augment the training data by generating a number $S = \{50, 100, 200\}$ of observation-action samples, for every timestep, by uniformly sampling states inside the tube. We denote the corresponding method as \ac{VSA}$-S$.

\subsection{Efficiency, robustness and performance}

\begin{figure}[t]
    \centering
    \includegraphics[trim={60 15 80 20},clip, width=\columnwidth]{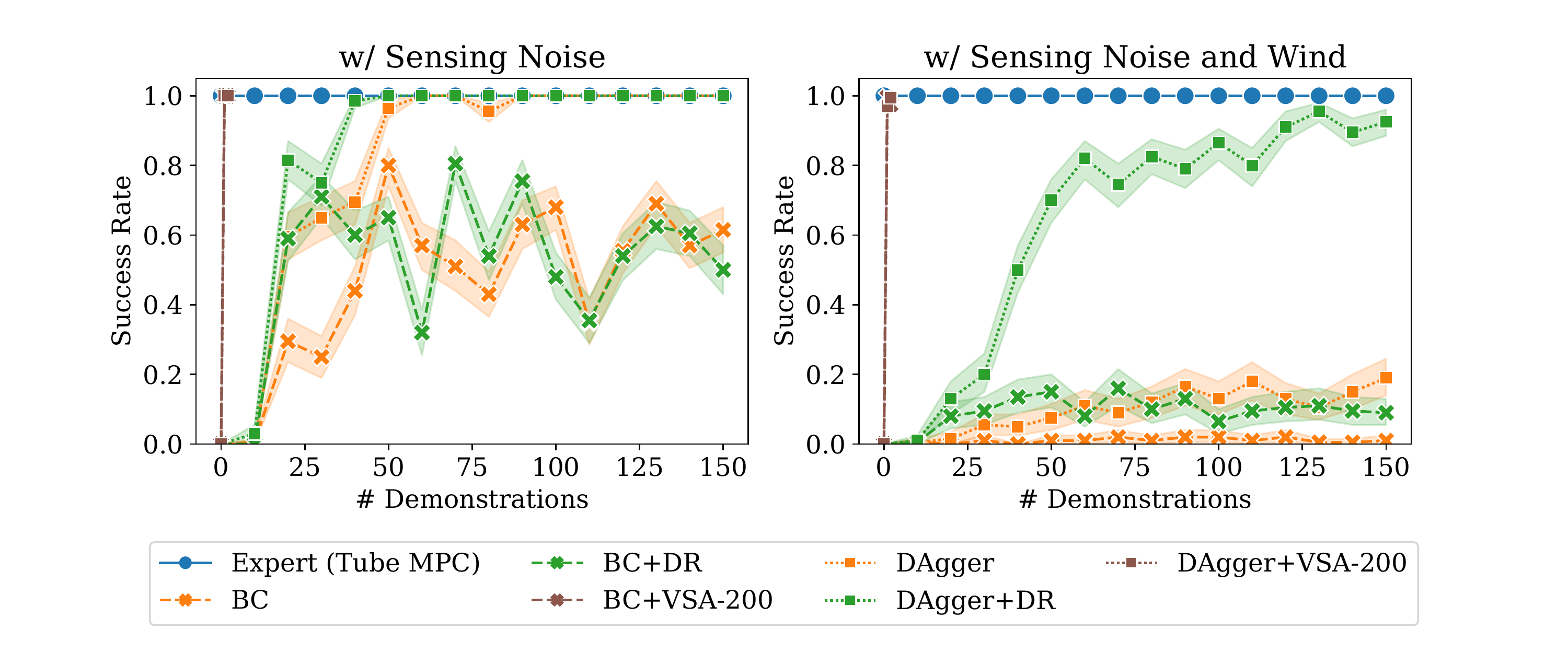}
    \caption{Robustness in the deployment domains under measurement uncertainties (sensing noise) and process noise (wind-like disturbances) as a function of the number of demonstrations. The shaded area represent the $95\%$ confidence intervals. Our approach, \acf{VSA}-$200$ (where $200$ indicates the number of samples extracted from the tube per timestep), after a single demonstration achieves full robustness in the environment w/o wind, and $>95\%$ robustness in the environment with wind. Evaluation performed across $10$ random seeds, with $20$ trajectories per seed, starting from different initial states. The lines of the \ac{VSA}-based approaches overlap.}
    \label{fig:one_traj_robustness}
    \vskip-4ex
\end{figure}

\begin{figure}[t]
    \centering
    \includegraphics[trim={60 5 80 20},clip,width=\columnwidth]{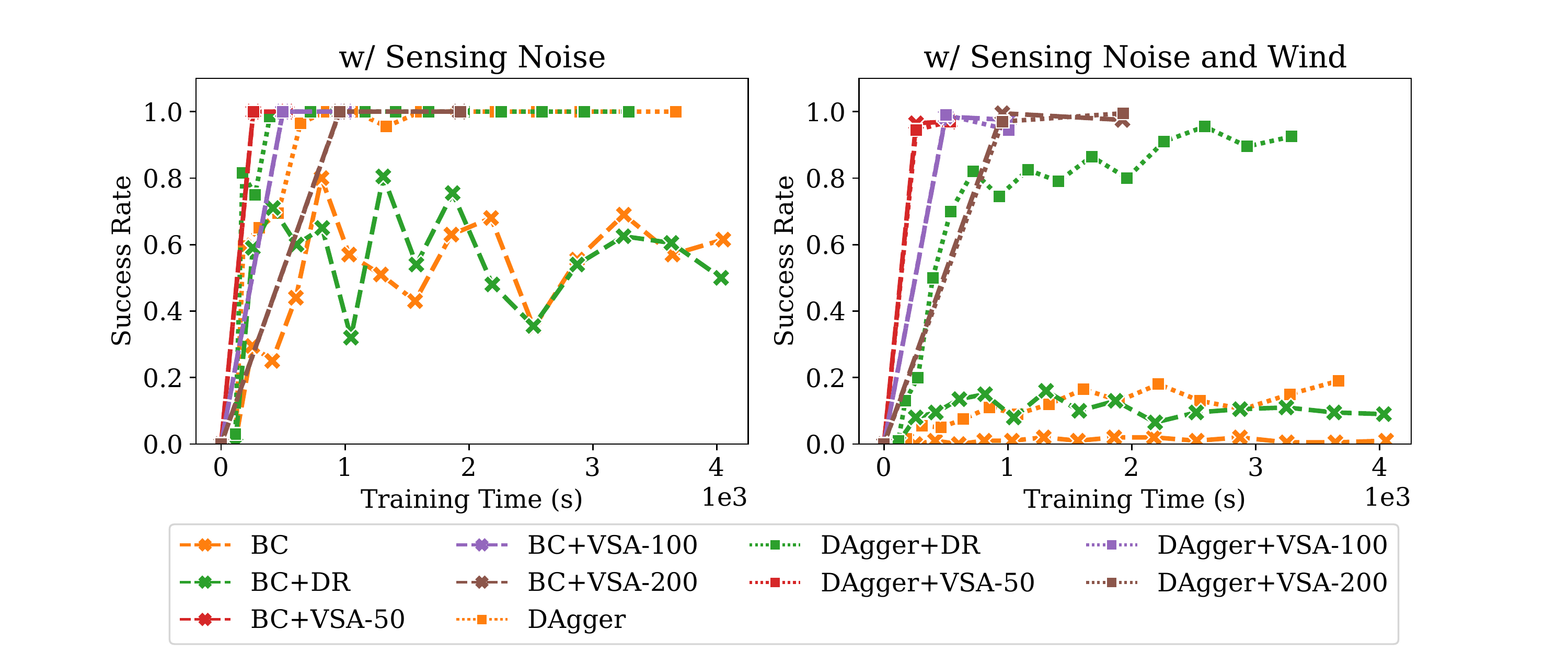}
    \caption{Robustness as a function of the training time, which includes the time to collect demonstrations from the expert and to train the policy. Robustness is considered under only measurement uncertainties (sensing noise) and in combination with process noise (wind-like disturbances). This result highlights that our approach, \acf{VSA}, is capable to achieve $> 95\%$ robustness under sensing \textit{and} process noise in less than half of the time than the best performing baseline approaches (DAgger+DR). Reducing the number of samples extracted from the tube (VSA-\{50, 100, 200\}) significantly reduces the training time, with a minimal change in robustness.}
    \label{fig:training_time}
    \vskip-5ex
\end{figure}

\begin{table}[t]
    \caption{Properties of the considered visuomotor policy learning approaches, including robustness, performance, and demonstration efficiency. We define an approach \textit{easy} if it does not require applying disturbances during the demonstration collection phase, and it is \textit{safe} if it does not violate state constraints (e.g., wall collision) during training. \textit{Performance} captures the relative error between the trajectory generated by output feedback RTMPC (expert) and the learned policy. \textit{Demonstration efficiency} indicates the number of demonstrations required for one approach to achieve at least $90\%$ \textit{success rate}. Performance and robustness of the baseline approaches are evaluated after $150$ demonstrations, while \ac{VSA} methods are evaluated after \textit{only} $2$ demonstrations.}
\newcolumntype{P}[1]{>{\centering\arraybackslash}p{#1}}
    \vskip-1ex
    \tiny
    \renewcommand{\tabcolsep}{1pt}
    \centering
    \begin{tabular}{|p{0.9cm}p{0.9cm}||P{0.7cm}|P{0.7cm}|P{0.7cm}|P{0.7cm}|P{0.7cm}|P{0.7cm}|P{0.7cm}|P{0.7cm}|}
    \hline
    \multicolumn{2}{|c||}{Method} &  
    \multicolumn{2}{c|}{Training} & 
    \multicolumn{2}{c|}{\makecell{Robustness\\succ. rate (\%, $\uparrow$)}} &
    \multicolumn{2}{c|}{\makecell{Performance\\expert gap (\%, $\downarrow$)}}  &
    \multicolumn{2}{c|}{\makecell{Demonstration\\Efficiency ($\downarrow$)}}\\
    
           Robustif.       & Imitation   & Easy          & Safe             & Noise                 &  Noise, Wind      &  Noise             &  Noise, Wind      & Noise             &  Noise, Wind \\
        \hline
        \hline
        \multirow{2}{*}{-} & BC          & \textbf{Yes}  & \textbf{Yes}     & 59.3                    &  0.8           & 98.4                  & 545.7      &     -             &  - \\
                           & DAgger      & \textbf{Yes}  & No               & \textbf{100.0}          &  17.0          & 8.3                   & 295.2              &     50            &   - \\ 
        \hline
        \multirow{2}{*}{DR}& BC         & No            & \textbf{Yes}      & 55.3                    &  9.3           & 128.0                  & 186.7              &     -             &  - \\ 
                        & DAgger        & No            & No                & \textbf{100.0}          &  91.0          & 5.4                   & 27.3              &     40            &  120 \\
        \hline
        \multirow{2}{*}{VSA-50} & BC   & \textbf{Yes}  & \textbf{Yes}      & \textbf{100.0}           & 97.0             & 27.4                 & 24.3              & \textbf{1}        & \textbf{1} \\
                            & DAgger   & \textbf{Yes}  & \textbf{Yes}      & \textbf{100.0}           & 97.0             & 13.6                 & 17.5              & \textbf{1}        & \textbf{1} \\
        \hline
        \multirow{2}{*}{VSA-100} & BC  & \textbf{Yes}  & \textbf{Yes}      & \textbf{100.0}          & 97.5             & 11.5                  & 14.1              & \textbf{1}        & \textbf{1} \\
                            & DAgger   & \textbf{Yes}  & \textbf{Yes}      & \textbf{100.0}          & 94.5             & 8.9                   & \textbf{13.0}              & \textbf{1}        & \textbf{1} \\
        \hline
        \multirow{2}{*}{VSA-200} & BC  & \textbf{Yes}  & \textbf{Yes}      & \textbf{100.0}          & 97.5             & 12.7                   & 17.9              & \textbf{1}        & \textbf{1} \\
                            & DAgger   & \textbf{Yes}  & \textbf{Yes}      & \textbf{100.0}          & \textbf{99.5}    & \textbf{4.5}           & 14.9              & \textbf{1}        & \textbf{1} \\
        \hline
    \hline
    \end{tabular}%
     \label{tab:analysis_at_convergence}
     \vskip-4ex
\end{table}

\noindent
\textbf{Robustness and efficiency.} We evaluate the robustness of the considered methods by monitoring the \textit{success rate}, as the percentage of episodes where the robot never violates any state constraint in the considered deployment environment. This metric is evaluated as a function of the number of demonstrations collected in the source environment.  The results are shown in \cref{fig:one_traj_robustness} and \cref{tab:analysis_at_convergence}, and highlight that all VSA methods, combined with either DAgger or BC, can achieve complete robustness in the environment without wind after a single demonstration, and at least $94.5\%$ success rate in the environment with wind (DAgger+VSA-$50$). A second demonstration can provide $> 99\%$ robustness even in the environment with wind (DAgger+VSA-$200$). The baseline approaches require instead $40$-$50$ demonstrations to achieve full success rate in the environment without wind, and the best performing baseline (DAgger+DR) requires $120$ demonstrations to achieve its top robustness ($97.3\%$ success rate) in the more challenging environment. We additionally evaluate the computational efficiency as the time (\textit{training time}) required to achieve a certain success rate. This is done by monitoring the robustness achieved at the $m$th iteration as a function of the time to reach that iteration, $\sum_{i=0}^{m}T^\text{iter}_i$. The results in \cref{fig:training_time} highlight that our approach requires less than \textit{half} training time than the best performing baseline (DAgger+DR) to achieve higher or comparable robustness in the most challenging environment. Additionally, reducing the number of samples in \ac{VSA} can enable large training time gains with a minimal loss in robustness.

\begin{figure}
\captionsetup[sub]{font=footnotesize}
\centering
\begin{subfigure}{1\columnwidth}
    \centering
    \includegraphics[trim={260, 73, 100, 65}, clip, width=0.85\columnwidth]{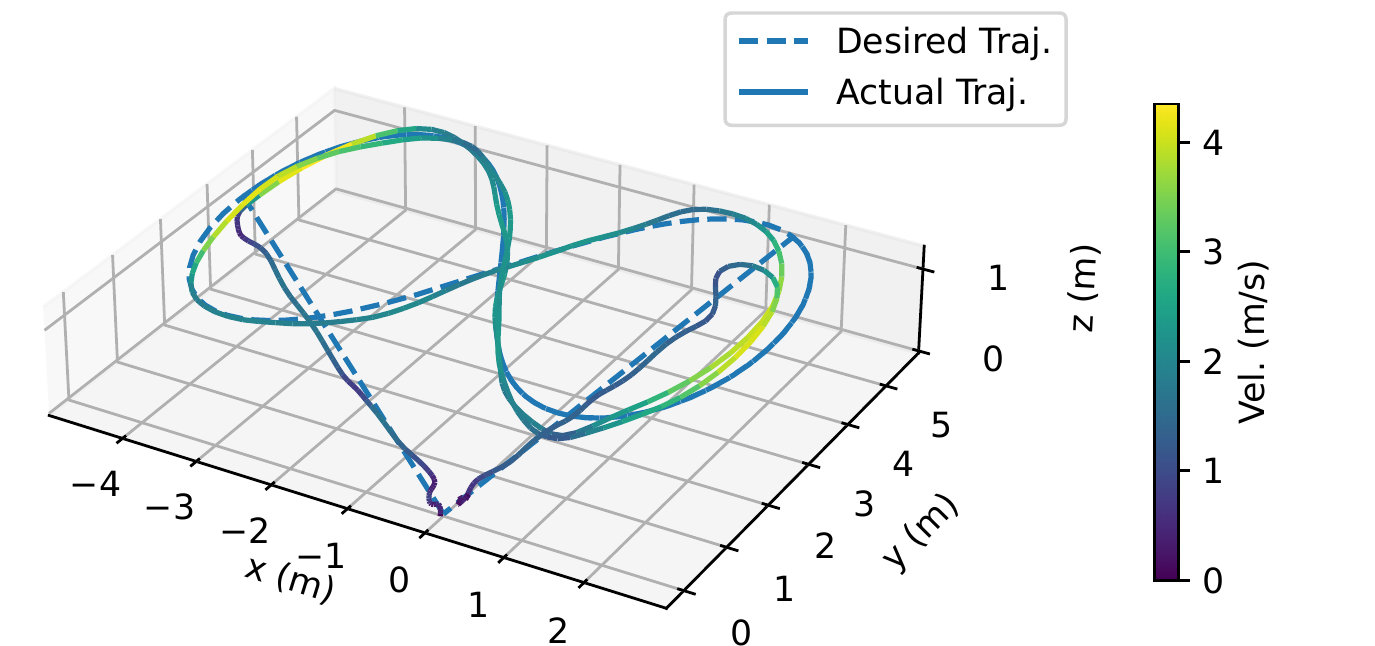}
    \caption{With sensing noise, no wind.}
    \label{fig:trajectory_no_wind}
\end{subfigure}%
\hspace{1.5cm}
\begin{subfigure}{1\columnwidth}
    \centering
    \includegraphics[trim={260, 73, 100, 65}, clip, width=0.85\columnwidth]{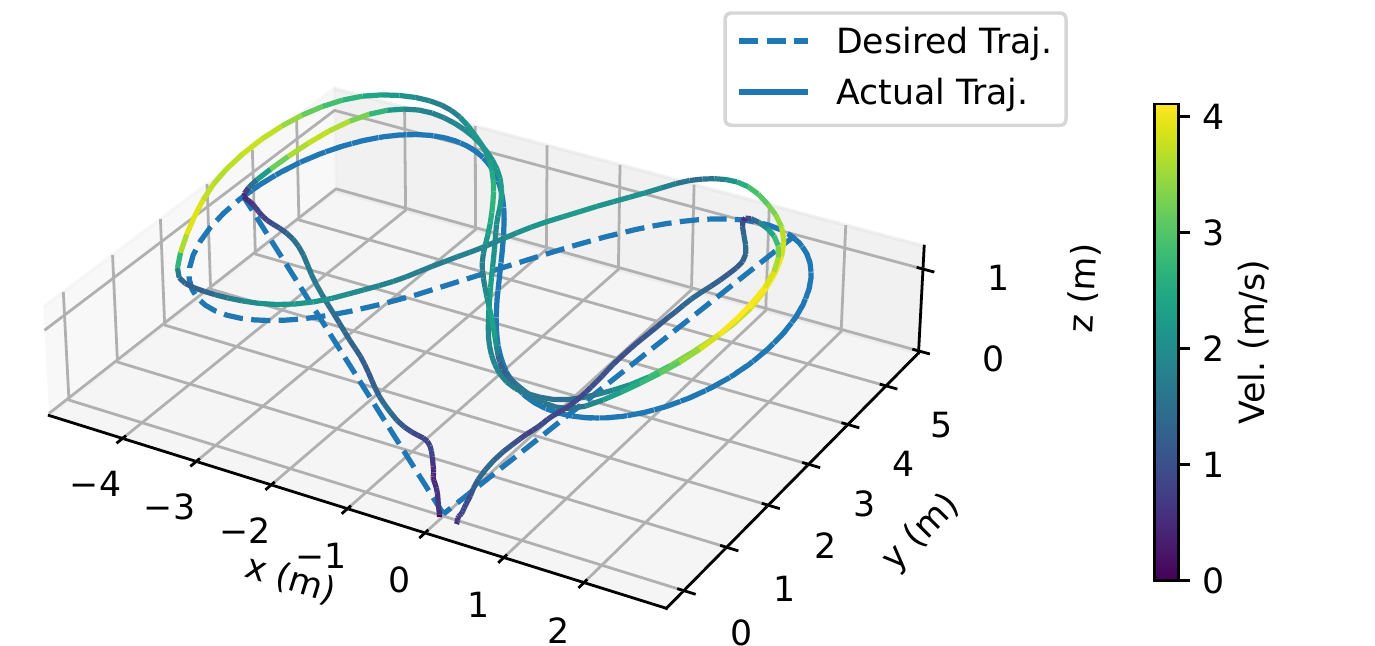}
    \caption{With sensing noise and wind gusts.}
    \vskip-1ex
    \label{fig:trajectory_no_wind}
\end{subfigure}
    \caption{Executed trajectory and velocity of the multirotor in our simulated environment under sensing noise and challenging wind-like disturbances. The policy has been trained from a \textit{single} demonstration using DAgger+\ac{VSA}-100. The wind disturbances applied are force steps, representing wind gusts, ranging $15\%$ to $20\%$ of the weight of the robot, and varying in magnitude and direction every $4.0$ s.}
    \label{fig:one_traj_performance}
    \vskip-2ex
\end{figure}

\begin{table}[]
\caption{Trajectory tracking errors. The position error is expressed in $m^2$, while velocity error in $m^2/s^2$.}
\label{tab:trajectory_mse}
\vskip-1ex
\centering
\begin{tabular}{@{}llcc|cc@{}}
\toprule
\multicolumn{2}{l}{\multirow{2}{*}{}}                                               & \multicolumn{2}{c|}{\textbf{DAgger+VSA-$100$}}                                         & \multicolumn{2}{c}{\textbf{Expert}}                                            \\ \cmidrule(l){3-6} 
\multicolumn{2}{l}{}                                                                & \multicolumn{1}{l|}{\textbf{Noise}} & \multicolumn{1}{l|}{\textbf{Noise, Wind}} & \multicolumn{1}{l|}{\textbf{Noise}} & \multicolumn{1}{l}{\textbf{Noise, Wind}} \\ \midrule
\multicolumn{1}{c|}{\multirow{2}{*}{\textbf{MSE}}} & \multicolumn{1}{l|}{Pos.}   & \multicolumn{1}{c|}{0.332}           & 0.593                                      & \multicolumn{1}{c|}{0.256}           & 0.562                                     \\ \cmidrule(l){2-6} 
\multicolumn{1}{c|}{}                              & \multicolumn{1}{l|}{Vel.} & \multicolumn{1}{c|}{0.579}           & 0.626                                      & \multicolumn{1}{c|}{0.563}           & 0.598                                     \\ \cmidrule(l){2-6} 
\end{tabular}

\vskip-5ex
\end{table}
\noindent
\textbf{Performance at convergence.} We evaluate the performance of the considered approaches via the \textit{expert gap}, capturing the relative error with respect to the output feedback RTMPC expert of the stage cost $\sum_{t} \mathbf \|\mathbf{e}_t\|_{\mathbf Q}^2+ \| \mathbf {u}_t \|^2_{\mathbf R}$ along the trajectories executed in the two considered deployment environment. We exclude from the performance evaluation those trajectories that violate state constraints (not robust).  \cref{tab:analysis_at_convergence} reports the  \textit{expert gap} at the convergence of the various methods. It highlights that DAgger combined with \ac{VSA} (DAgger+\ac{VSA}) performs comparably or better than the most robust baseline (DAgger+DR) but at a fraction of the number of demonstrations ($2$ instead of $150$) and training time. \cref{fig:one_traj_performance} shows a qualitative evaluation of the trajectory tracked by the learned policy and highlights that the robot remains close to the reference despite the relatively high velocity, wind, and sensing noise. \cref{tab:trajectory_mse} gives the corresponding MSE for the position and velocity, showing that the obtained visuomotor policy has tracking errors comparable to those of the output feedback \ac{RTMPC} expert, despite using vision information to infer the position of the robot, while the expert has access to noisy ground truth position information.

\noindent
\textbf{Efficiency at deployment.} \cref{tab:computational_cost} reports the time to compute a new action. It shows that on average the compressed policy is $17.6$ times faster, even when excluding the cost of state estimation, than the corresponding model-based controller. Evaluation performed on an Intel i9-10920 ($12$ cores) with two Nvidia RTX 3090 GPUs.
\begin{table}[t]
    \caption{Computation time required to generated a new action for the output feedback RTMPC (Expert) and the proposed learned visuomotor neural network policy (Policy), given a sensory observation. The learned neural network policy is $17.6$ times faster than the optimization-based expert.}
    \vskip-2ex
    \label{tab:computational_cost}    \centering
    \begin{tabular} {|C{1.0cm} |C{1.5cm}|C{0.9cm}|C{0.9cm}|C{0.9cm}|C{0.9cm}| }
    \multicolumn{2}{c}{} & \multicolumn{4} {c}{Time (ms)} \\
    \hline
        Method  & Setup & Mean & St. Dev. & Min & Max \\
        \hline 
        \hline 
        Expert            & CVXPY \cite{agrawal2018rewriting} &  $15.70$ & $13.37$ & $7.37$ & $76.56$  \\
        \textbf{Policy}   & PyTorch                       & $\mathbf{0.89}$ & $\mathbf{0.03}$ & $\mathbf{0.84}$ & $\mathbf{1.02}$  \\\hline
    \end{tabular}
\vskip-4ex
\end{table}

\section{CONCLUSIONS}
We have presented a strategy for efficient and robust sensorimotor policy learning. Our key idea relies on the co-design of controller and data augmentation strategy with \ac{IL} methods. Such controller and data augmentation approach leverage the theory of output feedback \ac{RTMPC} to compute relevant observations and actions for data augmentation---considering the effects of sensing noise, domain uncertainties, and disturbances. We have tailored our approach to the context of visuomotor policy learning, relying on a mesh of the considered environment to generate tube-guided augmented views. Our numerical evaluation has shown that the proposed method can learn a robust visuomotor policy from a few demonstrations, outperforming \ac{IL} baselines in demonstration and computational efficiency. 
Future works will focus on robustifying the policy to uncertainties/imperfections in the 3D mesh of the environment ($g_\text{cam}^{-1}$), accounting for state estimation dynamics in the data augmentation procedure, and performing hardware experiments.

\section*{ACKNOWLEDGMENT}
This work was funded by the Air Force Office of Scientific Research MURI FA9550-19-1-0386.

\balance
\bibliographystyle{IEEEtran}
\bibliography{root}
\begin{acronym}
\acro{OC}{Optimal Control}
\acro{LQR}{Linear Quadratic Regulator}
\acro{MAV}{Micro Aerial Vehicle}
\acro{GPS}{Guided Policy Search}
\acro{UAV}{Unmanned Aerial Vehicle}
\acro{MPC}{model predictive control}
\acro{RTMPC}{robust tube model predictive controller}
\acro{DNN}{deep neural network}
\acro{BC}{Behavior Cloning}
\acro{DR}{Domain Randomization}
\acro{SA}{Sampling Augmentation}
\acro{IL}{Imitation learning}
\acro{DAgger}{Dataset-Aggregation}
\acro{MDP}{Markov Decision Process}
\acro{VIO}{Visual-Inertial Odometry}
\acro{VSA}{Visuomotor Sampling Augmentation}
\end{acronym}

\end{document}